\documentclass[11pt,a4paper]{article}
\usepackage[hyperref]{naaclhlt2019}
\usepackage{times}
\usepackage{latexsym}

\usepackage{url}

\usepackage[pdftex]{graphicx}
\usepackage{amsmath}
\usepackage{bm}
\usepackage{capt-of}
\usepackage{amssymb}
\usepackage{adjustbox}
\usepackage{booktabs}
\usepackage{blindtext}
\usepackage{natbib}
\usepackage{multirow}

\usepackage{CJKutf8}

\aclfinalcopy 
\def\aclpaperid{773} 

\title{Segmentation-free Compositional $n$-gram Embedding}

\author{Geewook Kim \and Kazuki Fukui \and Hidetoshi Shimodaira \\
         Department of Systems Science, Graduate School of Informatics, Kyoto University\\
         Mathematical Statistics Team, RIKEN Center for Advanced Intelligence Project\\
         {\tt \{geewook, k.fukui\}@sys.i.kyoto-u.ac.jp, shimo@i.kyoto-u.ac.jp}
}

\date{}

\begin{document}
\maketitle
\begin{abstract}
We propose a new type of representation learning method that models words, phrases and sentences seamlessly.
Our method does not depend on word segmentation and any human-annotated resources (e.g., word dictionaries),
yet it is very effective for noisy corpora written in unsegmented languages such as Chinese and Japanese.
The main idea of our method is to ignore word boundaries completely (i.e., \textit{segmentation-free}), and construct representations for all character $n$-grams in a raw corpus with embeddings of \textit{compositional} sub-$n$-grams.
Although the idea is simple, our experiments on various benchmarks and real-world datasets show the efficacy of our proposal.
\end{abstract}

\section{Introduction}
Most existing word embedding models~\cite{Mikolov:2013:DRW:2999792.2999959,D14-1162,Q17-1010} take a sequence of words as their input.
Therefore, the conventional models are dependent on word segmentation \cite{P17-1078,Q18-1030}, which is a process of converting a raw corpus (i.e., a sequence of characters) into a sequence of segmented character $n$-grams.
After the segmentation, the segmented character $n$-grams are assumed to be words, 
and each word's representation is constructed from distribution of neighbour words that co-occur together across the estimated word boundaries.
However, in practice, this kind of approach has several problems.
First, word segmentation is difficult especially when texts in a corpus are noisy or unsegmented \cite{Saito2014MorphologicalAF, W18-6120}.
For example, word segmentation on social network service (SNS) corpora, such as Twitter, is a challenging task since it tends to include many misspellings, informal words, neologisms, and even emoticons.
This problem becomes more severe in unsegmented languages, such as Chinese and Japanese, whose word boundaries are not explicitly indicated.
Second, word segmentation has ambiguities~\cite{Luo:2002:CAR:1072228.1072283, Li:2003:UTO:1119250.1119251}.
For example, a compound word \begin{CJK}{UTF8}{min}線形代数学\end{CJK} (linear algebra) can be seen as a single word or sequence of words, such as \begin{CJK}{UTF8}{min}線形$|$代数学\end{CJK} (linear $|$ algebra).

Word segmentation errors negatively influence subsequent processes~\cite{xu-zens-ney:2004:SIGHAN}.
For example, we may lose some words in training corpora, leading to a larger Out-Of-Vocabulary (OOV) rate~\cite{I05-3001}.
Moreover, 
segmentation errors, such as segmenting \begin{CJK}{UTF8}{min}きのう\end{CJK} (yesterday) 
as \begin{CJK}{UTF8}{min}き$|$のう\end{CJK} (tree $|$ brain), produce false co-occurrence information.
This problem is crucial for most existing word embedding methods as they are based on distributional hypothesis~\cite{harris1954distributional}, which can be summarized as: ``\textit{a word is characterized by the company it keeps}''~\cite{firth1957}.

To enhance word segmentation, some recent works~\cite{jieba, sato2015mecabipadicneologd, niadic} made rich resources publicly available.
However, maintaining them up-to-date is difficult and it is infeasible for them to cover all types of words.
To avoid the negative impacts of word segmentation errors, \citet{D17-1080} proposed a word embedding method called \emph{segmentation-free word embedding} (\texttt{sembei}).
The key idea of \texttt{sembei} is to directly embed frequent character $n$-grams from a raw corpus without conducting word segmentation.
However, most of the frequent $n$-grams are non-words~\cite{W18-6120}, and hence \texttt{sembei} still suffers from the OOV problems. The fundamental problem also lies in its extension~\cite{W18-6120}, although it uses external resources to reduce the number of OOV.
To handle OOV problems, \citet{Q17-1010} proposed a novel \textit{compositional} word embedding method with subword modeling, called \emph{subword-information skip-gram} (\texttt{sisg}).
The key idea of \texttt{sisg} is to extend the notion of vocabulary to include subwords, namely, substrings of words, for enriching the representations of words by the embeddings of its subwords.
In \texttt{sisg}, the embeddings of OOV (or unseen) words are computed from the embedings of their subwords.
However, \texttt{sisg} requires word segmentation as a prepossessing step, and the way of collecting co-occurrence information is dependent on the results of explicit word segmentation.

\begin{figure}[!t]
    \centering
    \includegraphics[width=\linewidth, height=5.0cm]{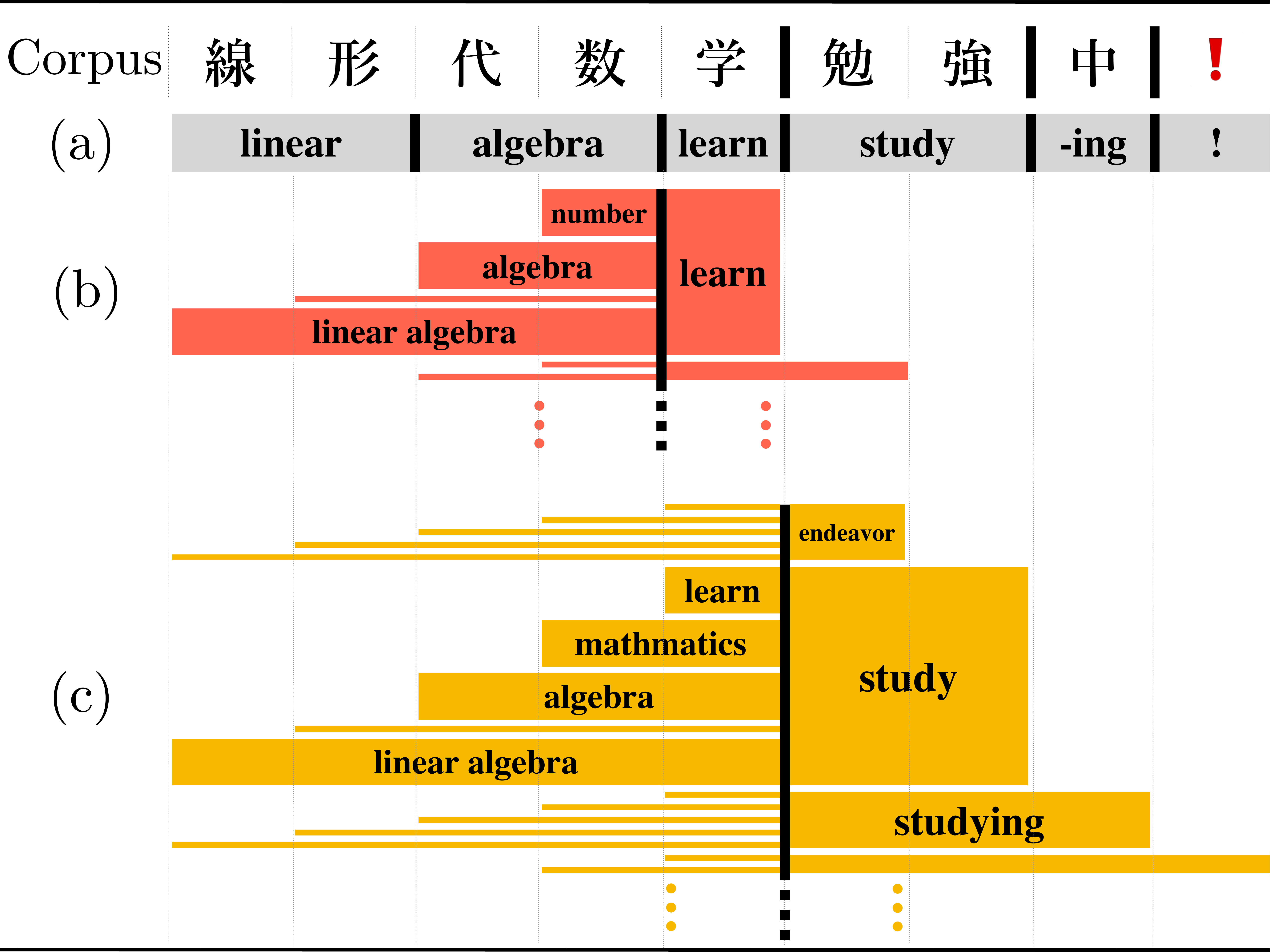}
    \captionof{figure}{A Japanese tweet with manual segmentation.
    (a) is the segmentation result of a widely-used word segmenter which conventional word embedding methods are dependent on.
    (b) and (c) show the embedding targets and their co-occurrence information to be considered in our proposed method \texttt{scne} on the boundaries of \begin{CJK}{UTF8}{min}数$|$学\end{CJK} and \begin{CJK}{UTF8}{min}学$|$勉\end{CJK}.
    Unlike conventional word embedding methods, \texttt{scne} considers all possible character $n$-grams on all boundaries (e.g., \begin{CJK}{UTF8}{min}線$|$形\end{CJK}, \begin{CJK}{UTF8}{min}形$|$代\end{CJK}, \begin{CJK}{UTF8}{min}代$|$数\end{CJK}, $\cdots$) in the raw corpus without segmentation.
    }
    \label{fig:1}
\end{figure}

For solving the issues of word segmentation and OOV, we propose a simple but effective unsupervised representation learning method for words, phrases and sentences, called \emph{segmentation-free compositional $n$-gram embedding} (\texttt{scne}).
The key idea of \texttt{scne} is to train embeddings of character $n$-grams to compose representations of all character $n$-grams in a raw corpus, and it enables treating all words, phrases and sentences seamlessly (see Figure~\ref{fig:1} for an illustrative explanation). 
Our experimental results on a range of datasets suggest that \texttt{scne} can compute high-quality representations for words and sentences although it does not consider any word boundaries and is not dependent on any human annotated resources.

\section{Segmentation-free Compositional $n$-gram Embedding (\texttt{scne})}

\begin{figure}[!t]
    \centering
    \includegraphics[width=\linewidth]{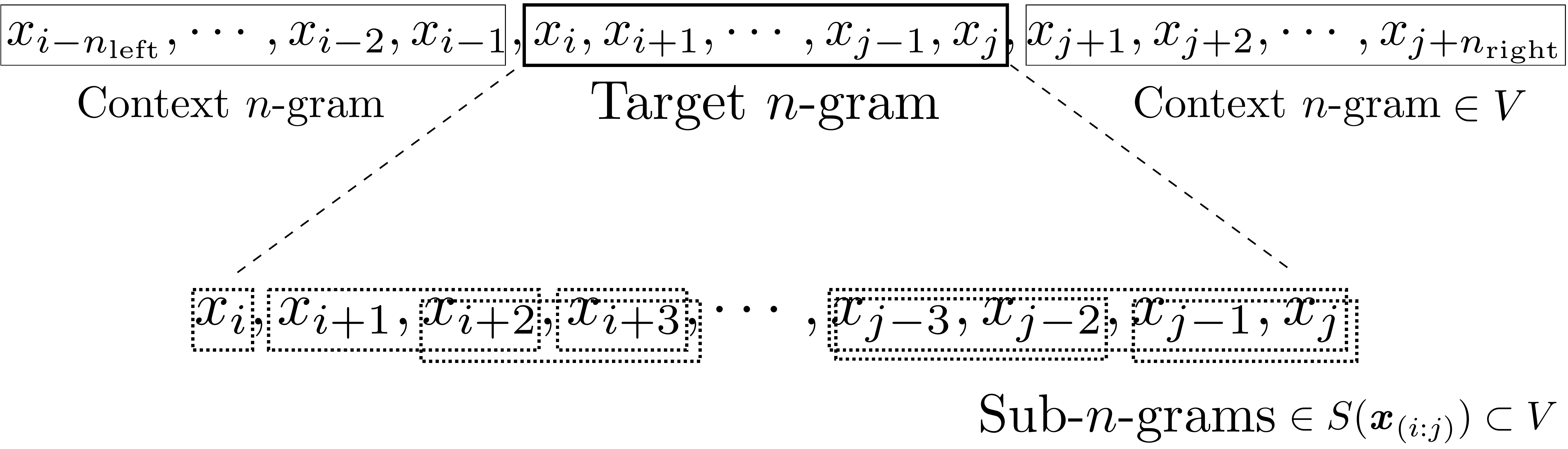}
    \captionof{figure}{A graphical illustration of the proposed model trying to compute a representation for a character $n$-gram $\bm{x}_{(i,j)}$. The co-occurrence of $\bm{x}_{(i,j)}$ and its neighbouring context $n$-grams are used to train embeddings of \textit{compositional} $n$-grams.}
    \label{fig:model}
\end{figure}

Our method \texttt{scne} successfully combines a subword model~\cite{Zhang:2015:CCN:2969239.2969312,D16-1157,Q17-1010,D18-1059}
with an idea of character $n$-gram embedding~\cite{D17-1080,W18-6120}.
In \texttt{scne}, the vector representation of a target character $n$-gram is defined as follows.
Let $x_1 x_2 \cdots x_N$ be a raw unsegmented corpus of $N$ characters.
For a range $i, i+1, \ldots, j$ specified by index $t=(i,j)$, $1\le i \le j \le N$, we denote the substring  $x_i x_{i+1} \cdots x_j$ as $\bm{x}_{(i,j)}$ or $\bm{x}_{t}$. 
In a training phase, \texttt{scne} first counts frequency of character $n$-grams in the raw corpus to construct $n$-gram set $V$ by collecting $M$-most frequent $n$-grams with $n \le n_{\rm{max}}$, where $M$ and $n_{\rm{max}}$ are hyperparameters.
For any target character $n$-gram $\bm{x}_{(i,j)}=x_i x_{i+1} \cdots x_j$ in the corpus,
\texttt{scne} constructs its representation $v_{\bm{x}_{(i,j)}}\in\mathbb{R}^d$ by summing the embeddings of its sub-$n$-grams as follows:
\[
 v_{\bm{x}_{(i,j)}} = \sum_{s \in S(\bm{x}_{(i,j)})} z_s, 
\]
where $S(\bm{x}_{(i,j)}) = \{ \bm{x}_{(i',j')} \in V \mid  i \le i' \le j' \le j \}$ consists of all sub-$n$-grams of target $\bm{x}_{(i,j)}$, and the embeddings of sub-$n$-grams $z_s\in\mathbb{R}^d$, $s\in V$ are model parameters to be learned.
The objective of \texttt{scne} is similar to that of \citet{Mikolov:2013:DRW:2999792.2999959},
\begin{align*}
  \scalebox{0.85}{$\displaystyle
  \sum_{t \in \mathcal{D}} \left\{
  \sum_{c \in \mathcal{C}(t)} \log\sigma \left(v^\top_{\bm{x}_t} u_{\bm{x}_c} \right) +
  \sum_{\tilde{s} \sim P_{\rm{neg}}}^k \log\sigma \left(- v^\top_{\bm{x}_t} u_{\tilde{s}} \right)
  \right\}
  $},
\end{align*}
where $\sigma(x)=\frac{1}{1+\exp(-x)}$, 
$\mathcal{D} = \{ (i,j) \mid 1\le i \le j \le N,  j-i+1 \le n_{\rm{target}} \}$, and $\mathcal{C}((i,j)) = \{ (i',j')  \mid \bm{x}_{(i',j')} \in V, j'=i-1 \text{ or } i'=j+1  \}$.
$\mathcal{D}$ is the set of indexes of all possible target $n$-grams in the raw corpus with $n\le n_{\rm{target}}$, where $n_{\rm{target}}$ is a hyperparameter.
$\mathcal{C}(t)$ is the set of indexes of contexts of the target $\bm{x}_{t}$, that is, all character $n$-grams in $V$ that are adjacent to the target (see Figures~\ref{fig:1} and \ref{fig:model}).
The negative sampling distribution $P_{\rm{neg}}$ of $\tilde s\in V$ is proportional to its frequency in the corpus.
The model parameters $z_s, u_{\tilde s} \in\mathbb{R}^d$, $s, \tilde s \in V$, are learned by maximizing the objective.
We set $n_{\rm{target}}=n_{\rm{max}}$ in our experiments.

Although we examine frequent $n$-grams for simplicity,
incorporating supervised word boundary information or \emph{byte pair encoding} into the construction of \emph{compositional} $n$-gram set would be an interesting future work~\cite{W18-6120,P16-1162,L18-1473}.

\begin{figure}[!t]
    \centering
    \includegraphics[width=\linewidth, height=4.0cm]{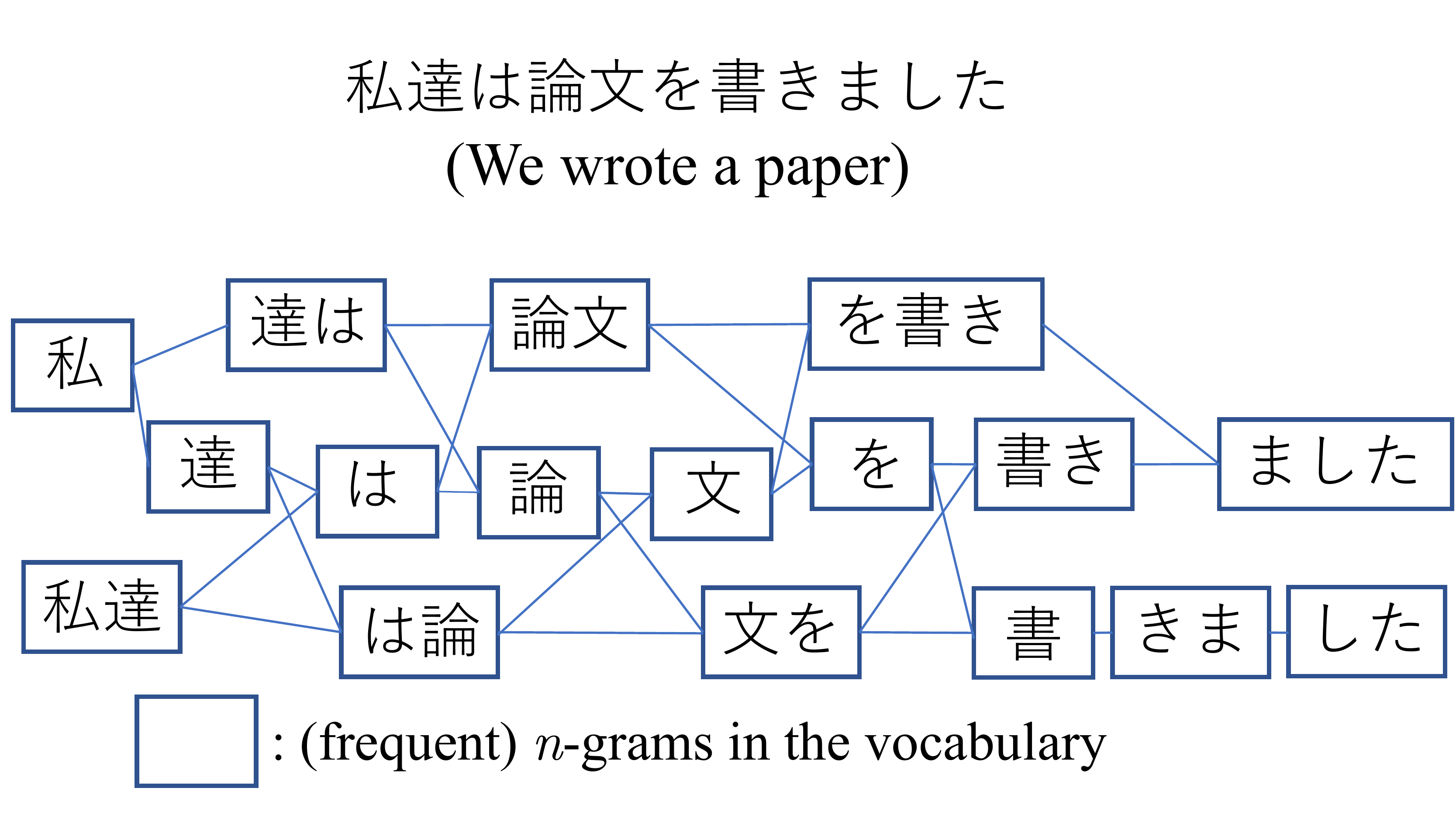}
    \caption{An example of a frequent $n$-gram lattice.}
    \label{fig:ngram_lattice}
\end{figure}

\subsection{Comparison to~\citet{D17-1080}} \label{sec:sembei}
To avoid the problems of word segmentation,
\citet{D17-1080} proposed \emph{segmentation-free word embedding} (\texttt{sembei})~\citep{D17-1080} that considers the $M$-most frequent character $n$-grams as individual words.
Then, a \textit{frequent $n$-gram lattice} is constructed, which is similar to a word lattice used in morphological analysis (see Figure~\ref{fig:ngram_lattice}).
Finally, the pairs of adjacent $n$-grams in the lattice are considered as target-context pairs and they are fed to existing word embedding methods, e.g., \texttt{skipgram}~\cite{Mikolov:2013:DRW:2999792.2999959}.
Although \texttt{sembei} is simple, the frequent $n$-gram vocabulary tends to include a vast amount of non-words~\citep{W18-6120}.
Furthermore, its vocabulary size is limited to $M$, hence, \texttt{sembei} can not avoid the undesirable issue of OOV. 
The proposed \texttt{scne} avoids these problems by taking all possible character $n$-grams as embedding targets.
Note that the target-context pairs of \texttt{sembei} are fully contained in those of \texttt{scne} (see Figure~\ref{fig:1}).

\subsection{Comparison to~\citet{W18-6120}}
To overcome the problem of OOV in \texttt{sembei}, \citet{W18-6120} proposed an extension of \texttt{sembei} called \emph{word-like $n$-gram embedding} (\texttt{wne}).
In \texttt{wne}, the $n$-gram vocabulary is filtered to have more vaild words by taking advantage of a supervised probabilistic word segmenter.
Although \texttt{wne} reduce the number of non-words,
there is still the problem of OOV since its vocabulary size is limited.
In addition, \texttt{wne} is dependent on word segmenter while \texttt{scne} does not.

\subsection{Comparison to~\citet{Q17-1010}}
To deal with OOV words as well as rare words,
\citet{Q17-1010} proposed \emph{subword information skip-gram} (\texttt{sisg}) that enriches word embeddings with the representations of its subwords, i.e., sub-character $n$-grams of words.
In \texttt{sisg}, a vector representation of a target word is encoded as the sum of the embeddings of its subwords.
For instance,
subwords of length $n = 3$ of the word \textit{where} are extracted as \texttt{<wh, whe, her, ere, re>}, where ``\texttt{<}'',``\texttt{>}'' are special symbols added to the original word to represent its left and right word boundaries. 
Then, a vector representation of \textit{where} is encoded as the sum of the embeddings of these subwords and that of the special sequence \texttt{<where>}, which corresponds to the original word itself.
Although \texttt{sisg} is powerful, it requires the information of word boundaries as its input, that is, semantic units need to be specified when encoding targets.
Therefore, it cannot be directly applied to unsegmented languages.
Unlike \texttt{sisg}, \texttt{scne} does not require such information.
The proposed \texttt{scne} is much simpler, but due to its simpleness,
the embedding target of \texttt{scne} should contains many non-words, which seems to be a problem (see Figure~\ref{fig:1}).
However, our experimental results show that \texttt{scne} successfully captures the semantics of words and even sentences for unsegmented languages without using any knowledge of word boundaries (see Section~\ref{sec:exp}).

\section{Experiments}\label{sec:exp}
In this section, we perform two intrinsic and two extrinsic tasks at both word and sentence level, focusing on unsegmented languages.
The implementation of our method is available on GitHub\footnote{\url{www.github.com/kdrl/SCNE}}.

\subsection{Common Settings}\label{sec:common}
\noindent\textbf{Baselines}:
We use \texttt{skipgram}~\cite{Mikolov:2013:DRW:2999792.2999959},
\texttt{sisg}~\cite{Q17-1010} and \texttt{sembei}~\cite{D17-1080} as word embedding baselines.
For sentence embedding, 
we first test simple baselines obtained by averaging the word vectors over a word-segmented sentence. 
In addition, 
we examine several recent successful sentence embedding methods,
\texttt{pv-dbow}, \texttt{pv-dm}~\cite{Le:2014:DRS:3044805.3045025} and \texttt{sent2vec}~\cite{pgj2017unsup} in an extrinsic task.
Note that both \texttt{scne} and \texttt{sembei} have embeddings of frequent character $n$-grams as their model parameters, 
but the differences come from training strategies, such as embedding targets and the way of collecting co-occurrence information (see Section~\ref{sec:sembei} for more details).
For contrasting \texttt{scne} with \texttt{sembei}, we also propose a variant of \texttt{sembei} (denoted by \texttt{sembei-sum}) as one of baselines, which composes word and sentence embeddings by simply summing up the embeddings of their sub-$n$-grams which are learned by \texttt{sembei}.

\noindent\textbf{Hyperparameters Tuning}:
To see the effect of rich resources for the segmentation-dependent baselines,
we employ widely-used word segmenter with two settings: Using only a basic dictionary ($\it{basic}$) or using a rich dictionary together ($\it{rich}$).
The dimension of embeddings is $200$, the number of epochs is $10$ and the number of negative samples is $10$ for all the methods.
The $n$-gram vocabulary size $M=2\times 10^6$ is used for \texttt{sisg}, \texttt{sembei} and \texttt{scne}.
The other hyperparameters, such as learning rate and $n_{\rm{max}}$, are carefully adjusted via a grid search in the validation set.
In the word similarity task,
$2$-fold cross validation is used for evaluation.
In the sentence similarity task,
we use the provided validation set.
In the downstream tasks, 
vector representations are combined with a supervised logistic regression classifier.
We repeat training and testing of the classifier $10$ times, 
while the prepared dataset is randomly split into train ($60\%$) and test ($40\%$) sets at each time,
and the hyperparameters are tuned by $3$-fold cross validation in the train set.
We adopt mean accuracy as the evaluation metric.
See Appendix~\ref{sec:appendix:exp} for more experimental details.

\begin{figure}[!t]
    \centering
    \includegraphics[width=\linewidth]{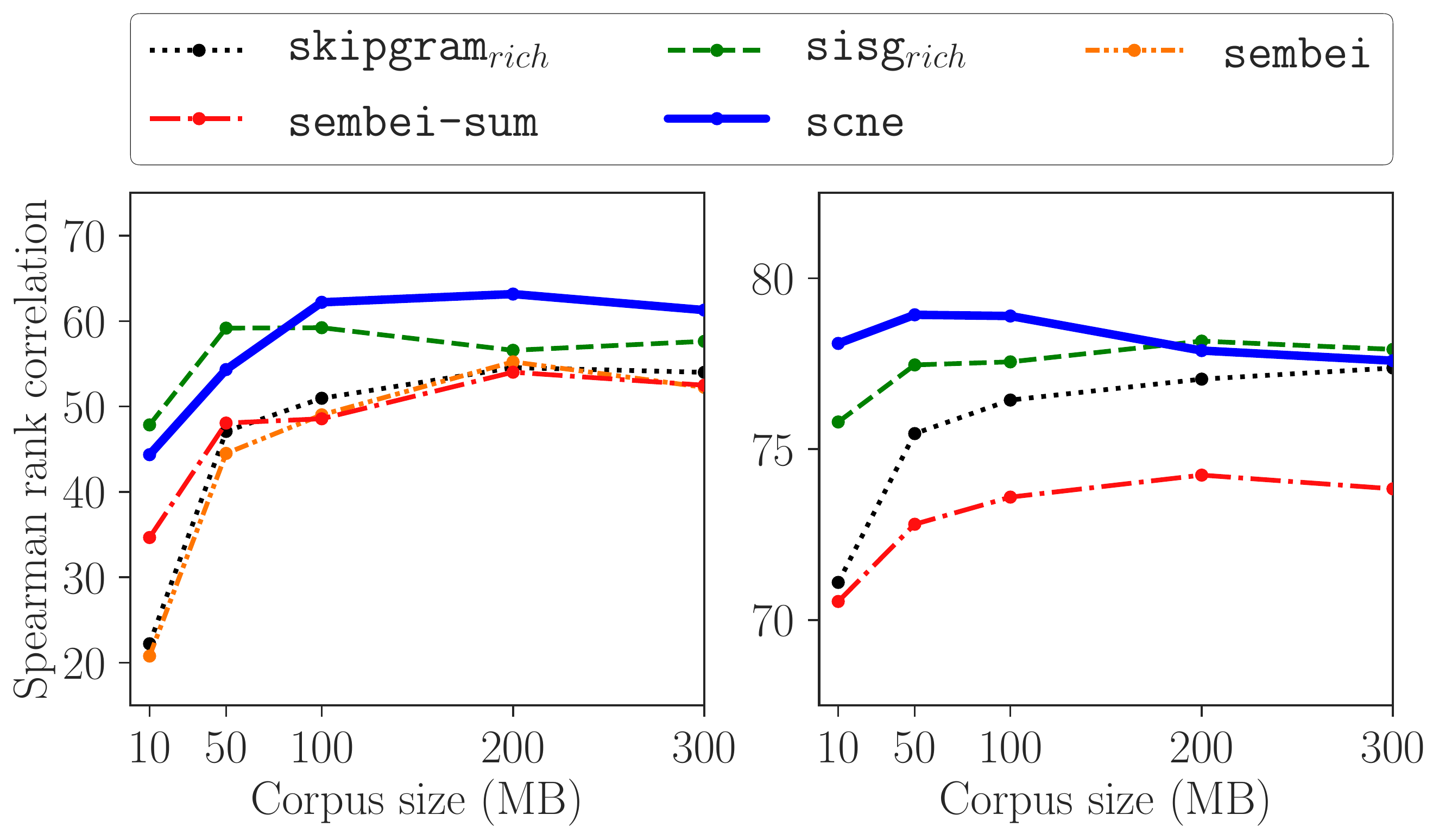}
    \captionof{figure}{Word (left) and sentence (right) similarity tasks on portions of Chinese Wikipedia corpus.}
    \label{fig:word-sim}
\end{figure}

\begin{table}[!t]
 \small
 \begin{center}
 \begin{adjustbox}{width=\linewidth}
 \begin{tabular}{lccccc}
  \toprule[0.2ex]
   & \texttt{skipgram}$_{rich}$ & \texttt{sisg}$_{rich}$ & \texttt{sembei} & \texttt{sembei-sum} & \textbf{\texttt{scne}}\\
  \midrule[0.1ex]
  Wiki.   & 51.0 & \underline{59.2} & 49.0 & 48.6  & \textbf{62.2}\\
  SNS & 41.3 & \underline{47.0} & 38.9 & 41.5  & \textbf{60.0}\\
  \cmidrule(lr){1-6}
  Diff. & -9.7 & -12.2 & -10.1 & -7.1 & -2.2\\
  \bottomrule[0.2ex]
 \end{tabular}
 \end{adjustbox}
 \caption{Spearman rank correlations of the word similarity task on two different Chinese corpora.
 Best scores are boldface and 2nd best scores are underlined.}
 \label{tab:noise}
 \end{center}
\end{table}

\begin{table*}[!t]
 \small
 \begin{center}
 \begin{adjustbox}{width=\textwidth}
 \begin{tabular}{lcccccccccccc}
\toprule[0.2ex]
                   & \multicolumn{6}{c}{Wikipedia corpora} & \multicolumn{6}{c}{Noisy SNS corpora}\\
                   \cmidrule(lr){2-7} \cmidrule(lr){8-13}
                   &\multicolumn{2}{c}{Chinese}&\multicolumn{2}{c}{Japanese}&\multicolumn{2}{c}{Korean}&\multicolumn{2}{c}{Chinese}&\multicolumn{2}{c}{Japanese}&\multicolumn{2}{c}{Korean}\\
                   \cmidrule(lr){2-3} \cmidrule(lr){4-5} \cmidrule(lr){6-7} \cmidrule(lr){8-9} \cmidrule(lr){10-11} \cmidrule(lr){12-13}
                   & \textbf{All} & Intersec. &  \textbf{All} & Intersec.  &  \textbf{All} & Intersec. &  \textbf{All} & Intersec. &  \textbf{All} & Intersec. &  \textbf{All} & Intersec. \\ 
\midrule[0.1ex]
\texttt{skipgram}$_{basic}$   &8.9  (11)               &81.0             &7.8  (10)                 &\underline{75.7}         &11.5 (15)               &\textbf{{77.1}}  &2.9  (5)               &58.2             &2.9  (7)               &41.4              & 2.4  (7)              &35.4   \\
\texttt{skipgram}$_{rich}$    &9.5  (12)               &81.0             &16.7 (20)                 &\textbf{{75.8}}          &11.9 (15)               &\underline{76.9} &3.0  (5)               &58.2             &4.1  (9)               &40.9              & 2.5  (7)              &34.2   \\
\texttt{sisg}$_{basic}$       &79.2 (100)              &\textbf{{82.3}}  &72.2 (100)                &\underline{75.7}         &72.2 (100)              &76.2             &71.0 (100)             &64.8             &67.1 (100)             &\underline{{46.9}}   & 63.4 (100)            &\textbf{{39.8}}   \\   
\texttt{sisg}$_{rich}$        &\underline{79.5} (100)  &\underline{82.2} &\underline{73.3} (100)    &74.7                     &\underline{72.4} (100)  &76.6             &70.8 (100)             &\underline{64.9} &\underline{67.5} (100) &46.0              & 63.3 (100)            &37.7   \\
\texttt{sembei}               &21.8 (25)               &79.0             &18.2 (23)                 &70.1                     &14.2 (19)               &41.8             &4.5  (7)               &59.6             &4.9  (10)              &41.9              & 5.0  (13)             &33.7  \\
\texttt{sembei-sum}           &76.8 (100)              &74.2             &69.9 (100)                &61.3                     &66.3 (100)              &56.0             &\underline{72.3} (100) &56.4             &66.3 (100)             &40.7              &\underline{64.3} (100) &34.8  \\
\textbf{\texttt{scne} (Proposed)}&\textbf{{79.8} (100)}&81.5             &\textbf{{73.9} (100)}     &74.0                     &\textbf{{73.2} (100)}   &73.9             &\textbf{{74.9} (100)}  &\textbf{{65.0}}  &\textbf{{68.1} (100)}  &\textbf{47.6}  &\textbf{{65.3} (100)}  &\underline{38.2}\\
\bottomrule[0.2ex]
 \end{tabular} 
 \end{adjustbox}
 \caption{Noun category prediction accuracies (higher is better) and coverages [$\%$] (in parentheses, higher is better).}
 \label{tab:noun-category}
 \end{center}
\end{table*}

\begin{table}[!t]
 \small
 \begin{center}
 \begin{adjustbox}{width=\linewidth}
 \begin{tabular}{lcccc}
  \toprule[0.2ex]
                                               & Segmentation-free     & Chinese         & Japanese       & Korean           \\
  \midrule[0.1ex]
  \texttt{pv-dbow}$_{basic}$                            &              &   82.84         &  85.24              &  84.16          \\
  \texttt{pv-dbow}$_{rich}$                             &              &   83.47         &  85.55              &  \underline{84.80}\\
  \texttt{pv-dm}$_{basic}$                              &              &   76.96         &  80.67              &  66.35          \\
  \texttt{pv-dm}$_{rich}$                               &              &   77.94         &  81.37              &  67.32          \\
  \texttt{sent2vec}$_{basic}$                           &              &   85.09         &  87.12              &  82.31          \\
  \texttt{sent2vec}$_{rich}$                            &              &   85.39         &  87.20              &  82.34          \\
  \texttt{skipgram}$_{basic}$                           &              &   85.79         &  86.76              &  84.06           \\
  \texttt{skipgram}$_{rich}$                            &              &   85.77         &  87.16              &  84.48           \\
  \texttt{sisg}$_{basic}$                               &              &   85.67         &  87.22              &  84.34          \\
  \texttt{sisg}$_{rich}$                                &              &   85.04         &  87.25  &  84.35          \\
  \texttt{sembei-sum}                                & $\checkmark$ &   83.41         &  80.80         &  74.98          \\
  \texttt{scne}$_{n_{\rm{max}} = 8}$                 & $\checkmark$ &   \underline{87.07}  &  \underline{87.42} &  84.15             \\
  \textbf{\texttt{scne}}$_{\bm{n_{\rm{max}} = 16}}$  & $\checkmark$ &   \textbf{{87.76}}&  \textbf{{88.03}}&  \textbf{{86.74}}    \\
  \bottomrule[0.2ex]
 \end{tabular}
 \end{adjustbox}
 \caption{Sentiment classification accuracies [$\%$].}
 \label{tab:sentiment-analysis}
 \end{center}
\end{table}

\subsection{Word and Sentence Similarity}
We measure the ability of models to capture semantic similarity for words and sentences in Chinese; see Appendix~\ref{sec:appendix:jp} for the experiment in Japanese.
Given a set of word pairs, or sentence pairs, and their human annotated similarity scores,
we calculated Spearman's rank correlation between the cosine similarities of the embeddings and the scores.
We use the dataset of \citet{S12-1049} and \citet{DBLP:conf/emnlp/WangZZ17} for Chinese word and sentence similarity respectively.
Note that the conventional models, such as \texttt{skipgram}, cannot provide the embeddings for OOV words,
while the \textit{compositional} models, such as \texttt{sisg} and \texttt{scne}, can compute the embeddings by using their subword modeling.
In order to show comparable results, we use the null vector for these OOV words following~\citet{Q17-1010}.

\noindent \textbf{Results}:
To see the effect of training corpus size,
we train all models on portions of Wikipedia\footnote{We use 10, 50, 100, 300MB of Wikipedia from the head.}.
The results are shown in Figure~\ref{fig:word-sim}.
As it can be seen, the proposed \texttt{scne} is competitive with or outperforms the baselines for both word and sentence similarity tasks.
Moreover, it is worth noting that \texttt{scne} provides high-quality representations even when the size of training corpus is small, which is crucial for practical real-world settings where rich data is not available.
For a next experiment to see the effect of noisiness of training corpus,
we test both noisy SNS corpus and the Wikipedia corpus\footnote{We use 100MB of Sina Weibo posts for Chinese SNS corpus and 100MB of Chinese Wikipedia corpus.} of the same size.
The results are reported in Table~\ref{tab:noise}.
As it can be seen,
the performance of segmentation-dependent methods (\texttt{skipgram}, \texttt{sisg}) are decreased greatly by the noisiness of the corpus, while \texttt{scne} degrades only marginally. 
The other two segmentation-free methods (\texttt{sembei}, \texttt{sembei-sum}) performed poorly.
This shows the efficacy of our method in the noisy texts. 
On the other hand, in preliminary experiments on English (not shown),
\texttt{scne} did not get better results than our segmentation-dependent baselines and it will be a future work to incorporate easily obtainable word boundary information into \texttt{scne} for segmented languages.

\subsection{Noun Category Prediction}
As a word-level downstream task, we conduct a noun category prediction on Chinese, Japanese and Korean\footnote{Although Korean has spacing, word boundaries are not obviously determined by space.}.
Most settings are the same as those of~\citet{D17-1080}.
Noun words and their semantic categories are extracted from Wikidata~\cite{42240} with a predetermined semantic category set\footnote{\{food, song, music band name, manga, fictional character name, television series, drama, chemical compound, disease, taxon, city, island, country, year, business enterprise, public company, profession, university, language, book\}},
and the classifier is trained to predict the semantic category of words from the learned word representations,
where unseen words are skipped in training and treated as errors in testing.
To see the effect of the noisiness of corpora,
both noisy SNS corpus and Wikipedia corpus of the same size are examined as training corpora\footnote{For each language, we use 100MB of Wikipedia and SNS data as training corpora. For the SNS data, we use Sina Weibo for Chinese and Twitter for the rest.}.

\noindent \textbf{Results}:
The results are reported in Table~\ref{tab:noun-category}.
Since the set of covered nouns (i.e., non-OOV words) depends on the methods,
we calculate accuracies in two ways for a fair comparison:
Using all the nouns and using the intersection of the covered nouns.
\texttt{scne} achieved the highest accuracies in all the settings when using all the nouns,
and also performed well when using the intersection of the covered nouns,
especially for the noisy corpora.

\subsection{Sentiment Analysis}
As a sentence-level evaluation,
we perform sentiment analysis on movie review data.
We use 101k, 56k and 200k movie reviews and their scores respectively from Chinese, Japanese and Korean movie review websites (see Appendix~\ref{sec:appendix:mr} for more details).
Each review is labeled as positive or negative by its rating score.
Sentence embedding models are trained using the whole movie reviews as training corpus.
Among the reviews, 5k positive and 5k negative reviews are randomly selected, 
and the selected reviews are used to train and test the classifiers as explained in Section~\ref{sec:common}.

\noindent \textbf{Results}:
The results are reported in Table~\ref{tab:sentiment-analysis}.
The accuracies show that \texttt{scne} is also very effective in the sentence-level application.
In this experiment, we observe that the larger $n_{\rm{max}}$ contributes to the performance improvement in sentence-level application by allowing our model to capture composed representations for longer phrases or sentences.

\section{Conclusion}
We proposed a simple yet effective unsupervised method to acquire general-purpose vector representations of words, phrases and sentences seamlessly,
which is especially useful for languages whose word boundaries are not obvious, i.e., unsegmented languages.
Although our method does not rely on any manually annotated resources or word segmenter,
our extensive experiments show that our method outperforms the conventional approaches
that depend on such resources.

\section*{Acknowledgments}
We would like to thank anonymous reviewers for their helpful advice.
This work was partially supported by JSPS KAKENHI grant 16H02789 to HS and 18J15053 to KF.

\bibliographystyle{acl_natbib}

\appendix

\section{Appendices}
\subsection{Experimental Details} \label{sec:appendix:exp}
\subsubsection{Hyperparameters Tuning} \label{sec:appendix:tuning}
For \texttt{skipgram}, we performed a grid search over $(h, \gamma) \in \{1, 5, 10\} \times \{0.01, 0.025\}$, where $h$ is the size of context window and $\gamma$ is the initial learning rate.
For \texttt{sisg}, we performed a grid search over $(h, \gamma, n_{\rm{min}}, n_{\rm{max}}) \in \{1, 5, 10\} \times \{0.01, 0.025\} \times \{1, 3\} \times \{4, 8, 12\}$, where $h$ is the size of context window, $\gamma$ is the initial learning rate, $n_{min}$ is the minimum length of character $n$-gram and $n_{max}$ is the maximum length of character $n$-gram.
For \texttt{pv-dbow}, \texttt{pv-dm} and \texttt{sent2vec}, we performed a grid search over $(h, \gamma) \in \{5, 10\} \times \{0.01, 0.05, 0.1, 0.2, 0.5\}$, where $h$ is the size of context window and $\gamma$ is the initial learning rate.
For \texttt{sembei} and \texttt{scne}, 
we used the initial learning rate $0.01$ and $n_{\rm{min}}=1$.
The maximum length of $n$-gram to consider $n_{\rm{max}}$ is grid searched over $\{4,6,8\}$ 
in the word and sentence similarity tasks.
In the noun category prediction task, we used $n_{\rm{max}}=8$ for \texttt{sembei} and the $n_{\rm{max}}$ of \texttt{scne} is grid searched over $\{4,6,8\}$.
For sentiment analysis task, 
we tested both $n_{\rm{max}}=8$ and $n_{\rm{max}}=16$ for \texttt{sembei} and \texttt{scne} 
to see the effect of large $n_{\rm{max}}$.
After carefully monitoring the loss curve and the performance in the word and sentence similarity tasks,
we set the number of epochs $10$ for all methods.
In preliminary experiments, we also tested the number of epochs $20$ for the word-segmentation-dependent baselines but there were no significant differences.
In the two supervised downstream tasks,
the learned vector representations are combined with the logistic regression classifier.
The parameter $C$, which is the inverse of regularization strength of the classifier, 
is adjusted via a grid search over $C \in \{0.1, 0.5, 1, 5, 10\}$.
Again, as explained in the main paper, the hyperparamters are grid searched on the determined validation set for all experiments.

\subsubsection{Implementations}
Here we provide the list of implementations of baselines which are used in our experiments.
For \texttt{skipgram}\footnote{\url{https://code.google.com/archive/p/word2vec/}},
\texttt{sisg}\footnote{\url{https://github.com/facebookresearch/fastText}},
\texttt{sembei}\footnote{\url{https://github.com/oshikiri/w2v-sembei}},
and \texttt{sent2vec}\footnote{\url{https://github.com/epfml/sent2vec}},
we use the official implementations provided by the authors.
Meanwhile, as for \texttt{pv-dbow} and \texttt{pv-dm}, we employ a widely-used implementation of Gensim library\footnote{\url{https://radimrehurek.com/gensim/models/doc2vec.html}}.

\subsubsection{Word Segmenters and Word Dictionaries for Unsegmented Languages}
Below we list the word segmentation tools and word dictionaries which are used in our experiments.
We employed a widely-used word segmentation tool for each language.

For Chinese language, 
we used jieba\footnote{\url{https://github.com/fxsjy/jieba}} with its default dictionary\footnote{\url{https://github.com/fxsjy/jieba/blob/master/jieba/dict.txt}} or with an extended dictionary\footnote{\url{https://github.com/fxsjy/jieba/blob/master/extra_dict/dict.txt.big}}, which fully supports both traditional and simplified Chinese characters.

For Japanese, we used MeCab\footnote{\url{http://taku910.github.io/mecab/}\label{foot9}} with its default dictionary called IPADIC$^\text{\ref{foot9}}$ along with specially designed neologisms-extended dictionary called mecab-ipadic-NEologd\footnote{\url{https://github.com/neologd/mecab-ipadic-neologd}}.
Note that, because this extended dictionary \emph{mecab-ipadic-NEologd} is specially designed to include many neologisms,
there is a significant word coverage improvement by using this word dictionary 
as it can be seen in the Japanese noun category prediction task in the main paper.

For Korean, we used mecab-ko\footnote{\url{https://bitbucket.org/eunjeon/mecab-ko}} with its default dictionary called mecab-ko-dic\footnote{\url{https://bitbucket.org/eunjeon/mecab-ko-dic}} along with another extended dictionary called NIADic\footnote{\url{https://github.com/haven-jeon/NIADic}}.

\subsubsection{Training Corpora}
We prepared Wikipedia corpora and SNS corpora for Chinese, Japanese and Korean for our experiments.
For the Wikipedia corpora,
we used the first 10, 50, 100, 200 and 300MB of texts from the publicly available Wikipedia dumps\footnote{\url{https://dumps.wikimedia.org/}}.
The texts are extracted by using WikiExtractor tool\footnote{\url{https://github.com/attardi/wikiextractor}}.
For Chinese SNS corpus,
we used 100MB of Leiden Weibo Corpus \cite{van2012leidon} from the head.
For Japanese and Korean SNS corpora,
we collected Japanese and Korean tweets using Twitter Streaming API.
We removed usernames and URLs from the SNS corpora.
There were many informal words, emoticons and misspellings in the SNS corpora.
We preserved them without preprocessing to see the effect of the noisiness of training corpora in our experiments.

\subsubsection{Preprocess of Wikidata}
For the noun category prediction task,
we extracted noun words and their semantic categories from Wikidata~\cite{42240} following \citet{D17-1080}.
We determined the semantic category set used in our experiments as follows:
First, we collected Wikidata objects that have Chinese, Japanese, Korean and English labels.
Next, we sorted the categories by the number of noun words, and removed categories (e.g., \textit{Wikimedia category} or \textit{Wikimedia template}) that do not represent any semantic category. We also removed out several categories that contain too many noun words (e.g., \textit{human}) or too few noun words (e.g., \textit{academic discipline}).
Since there were several duplicated labels for different Wikidata objects, the number of nouns for each language is slightly different.
Each category has at least 0.1k words and no more than 5k words.
The numbers of extracted noun words that are used in our experiments were 22,468, 22,396 and 22,298 for Chinese, Japanese and Korean, respectively.

\subsubsection{Movie Review Datasets} \label{sec:appendix:mr}
In the main paper, three movie review datasets are used to evaluate the quality of sentence embeddings.
We used 101,114, 55,837 and 200,000 movie reviews and their rating scores from Yahoo\begin{CJK}{UTF8}{min}奇摩電影\end{CJK}\footnote{\url{https://github.com/fychao/ChineseMovieReviews}},
Yahoo!\begin{CJK}{UTF8}{min}映画\end{CJK}\footnote{\url{https://github.com/dennybritz/sentiment-analysis/tree/master/data}}
and Naver Movies\footnote{\url{https://github.com/e9t/nsmc}} for Chinese, Japanese and Korean, respectively.

\begin{table}[!t]
 \small
 \begin{center}
 \begin{adjustbox}{width=0.475\textwidth}
 \begin{tabular}{lccccc}
  \toprule[0.2ex]
  & \texttt{skipgram}$_{rich}$ & \texttt{sisg}$_{rich}$ & \texttt{sembei} & \texttt{sembei-sum} & \textbf{\texttt{scne}}\\
  \midrule[0.1ex]
  Wiki. & 8.3 & \underline{15.4} & 4.0 & 9.3 &  \textbf{24.1}\\
  SNS   & 5.3 & \underline{12.7} & 2.8 & 9.3 & \textbf{23.0}\\
  \cmidrule(lr){1-6}
  Diff. & -3.0 & -2.7 & -1.2 &  -0.0 & -1.1\\
  \bottomrule[0.2ex]
 \end{tabular}
 \end{adjustbox}
 \caption{Spearman rank correlations of the word similarity task on two different Japanese corpora.}
 \label{tab:noise_jap}
 \end{center}
\end{table}

\subsection{Additional Experiment on Japanese} \label{sec:appendix:jp}
In this section, we show the results of Japanese word similarity experiments.
We use the datasets of \citet{DBLP:journals/corr/SakaizawaK17}.
It contains 4427 pairs of words with human similarity scores.
We omit sentence similarity task since there is no public widely-used benchmark dataset for Japanese yet.
Following the main paper, 
given a set of word pairs and their human annotated similarity scores,
we calculated Spearman's rank correlation between the cosine similarities of the embeddings and the human scores.
We use 2-fold cross validation for hyperparameters tuning.
The same grid search is performed as explained in Section~\ref{sec:appendix:tuning}.
To see the effect of the noisiness of training corpora,
we use two Japanese corpora, 100MB of Wikipedia corpus and 100MB of noisy SNS corpus (Twitter), which are also used in the Japanese noun category prediction task in the main paper.
As seen in Table~\ref{tab:noise_jap}, the experiment results for Japanese are similar to those of Chinese in the main paper.

\end{document}